\pdfoutput=1

\documentclass[11pt]{article}

\usepackage[preprint]{acl}

\usepackage{times}
\usepackage{latexsym}

\usepackage[T1]{fontenc}

\usepackage[utf8]{inputenc}

\usepackage{microtype}

\usepackage{inconsolata}

\usepackage{graphicx}
\usepackage{amsmath}
\usepackage{amsfonts}
\usepackage{booktabs}
\usepackage{caption}
\usepackage{subcaption}
\usepackage{makecell}
\usepackage{longtable}
\usepackage{tabularx}
\usepackage{supertabular}
\usepackage{afterpage}
\usepackage{multirow}

\title{Toward Automatic Safe Driving Instruction:\\A Large-Scale Vision Language Model Approach}

\author{
  \textbf{Haruki Sakajo\textsuperscript{1}},
  \textbf{Hiroshi Takato\textsuperscript{2}}
  \textbf{Hiroshi Tsutsui\textsuperscript{2}}
  \textbf{Komei Soda\textsuperscript{2,3}}\\
  \textbf{Hidetaka Kamigaito\textsuperscript{1}}, 
  \textbf{Taro Watanabe\textsuperscript{1}}
\\
  \textsuperscript{1}Nara Institute of Science and Technology (NAIST),\\\textsuperscript{2}Teatis inc., \textsuperscript{3}Queensland university of technology
\\
\texttt{sakajo.haruki.sd9@naist.ac.jp}\\
\texttt{\{tak, james\}@dotsfty.com}
\texttt{komei.soda@connect.qut.edu.au}\\
\texttt{\{kamigaito.h, taro\}@is.naist.jp}
}

\begin{document}

\maketitle
\begin{abstract}
Large-scale Vision Language Models (LVLMs) exhibit advanced capabilities in tasks that require visual information, including object detection.
These capabilities have promising applications in various industrial domains, such as autonomous driving.
For example, LVLMs can generate safety-oriented descriptions of videos captured by road-facing cameras.
However, ensuring comprehensive safety requires additional monitoring driver-facing views to detect risky events, such as the use of mobiles while driving.
Thus, the ability to process synchronized inputs is necessary from both driver-facing and road-facing cameras.
In this study, we develop a model integrating two video inputs and investigate the capabilities of LVLMs by constructing a dataset and evaluating their performance on this dataset.
Our experimental results demonstrate that while pre-trained LVLMs have limited effectiveness, fine-tuned LVLMs can generate accurate and safety-aware driving instructions.
Nonetheless, several challenges remain, particularly in detecting subtle or complex events in the video.
Our findings and error analysis provide valuable insights that can contribute to the improvement of LVLM-based systems in this domain.
\end{abstract}

\section{Introduction}
The promising capabilities of Large Language Models (LLMs) are changing this society by assisting various tasks, e.g., coding~\cite{rozière2024codellamaopenfoundation} and education~\cite{liu-etal-teaching}.
Large-scale Vision Language Models (LVLMs) possess high capabilities in the intersection of vision and language tasks, leveraging the capabilities of LLMs, such as inference and instruction following, by integrating a vision encoder.
Therefore, LVLMs have been adopted across domains that require both visual and textual information, including the medical application~\cite{NEURIPS2023_5abcdf8e, yan-etal-2024-worse, pal-sankarasubbu-2024-gemini} and driving assistance~\cite{covla_wacv2025, Duan_2024_CVPR, Xuan_2024_CVPR}.

In a driving domain, LVLMs are used to detect objects on the road, generate actions, and provide safe driving instructions~\cite{zhou-etal-vlm-ar, Lu_Yao_Tu_Shao_Ma_Zhu_2025, Duan_2024_CVPR, Xuan_2024_CVPR}.
With the rapid growth of the dashcam industry, LVLMs are increasingly exploited to interpret driving scenes captured by the dashcam.
Previous studies investigated the capabilities of LVLMs to interpret vehicle behavior and suggest actions for safe driving~\cite{Duan_2024_CVPR, Xuan_2024_CVPR}.
However, while a single dashcam for the road-facing view is enough to detect risky actions of vehicles (e.g., harsh turns), a driver-facing view using an additional dashcam is needed, for example, when detecting the driver's use of mobiles while driving.

\begin{figure}[t]
    \centering
    \includegraphics[width=\linewidth]{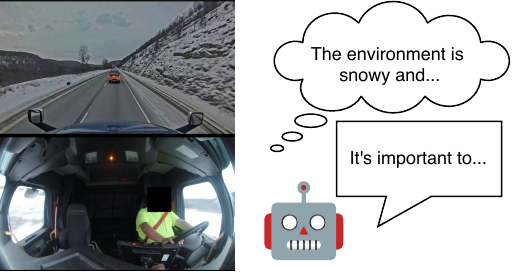}
    \caption{Illustration of an application of this study. A Model provides driving instructions for the given video.}
    \label{fig:overview}
\end{figure}

\begin{figure*}[t]
    \centering
    \includegraphics[width=\linewidth]{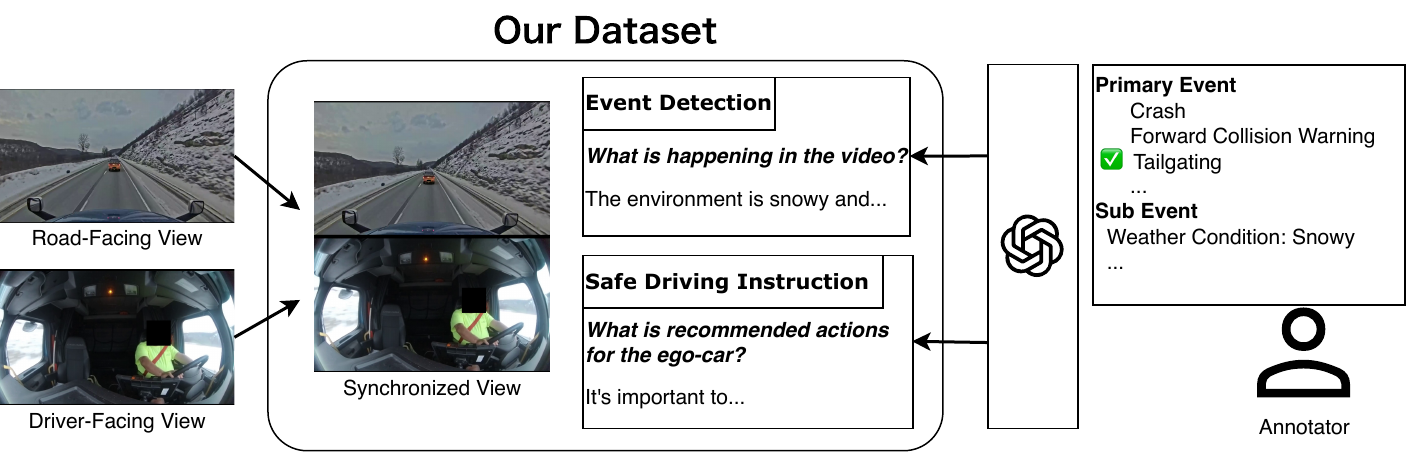}
    \caption{Overview of our dataset construction approach. The dataset contains videos with synchronized driver-facing and road-facing views. GPT-4o generates the gold answers to event detection and safe driving instruction questions based on annotated labels for each video.}
    \label{fig:dataset}
\end{figure*}

\begin{table*}[t]
    \centering
    \small
    \begin{tabular}{@{}lp{14cm}@{}}
    \toprule
          Human & What is happening in the video? \\
          Assistant & The environment is icy with clear weather as you approach an intersection. A rolling stop occurs as you navigate the intersection at 8 mph while properly using your turn signal.\\
          \midrule
          Human & What is recommended actions for the ego-car? \\
          Assistant &  \textbackslash ud83d\textbackslash udca1Paying extra attention to road conditions due to ice is important. Make sure to bring your vehicle to a complete stop at intersections to ensure safety, especially in icy conditions where stopping distances are longer. \\
    \bottomrule
    \end{tabular}
    \caption{An example of the question and its answer. The top is the event detection, and the bottom is the safe driving instruction.}
    \label{tab:qa_example}
\end{table*}

In this study, we develop LVLM-based models that generate instructions for safe driving based on two video streams that synchronize driver-facing and road-facing views.
We collect such videos and annotate each clip in two stages: first, by detecting events that happened in them, and second, by generating safe driving instructions.
We create a conversation-style dataset, where a user asks what is happening in the video, followed by a request to generate safe driving instructions.
Our experiments demonstrate that fine-tuned LVLMs on our dataset can generate safe driving instructions that are well-aligned with the visual context, whereas pre-trained models generate generic guidance.
Nonetheless, even the fine-tuned models occasionally fail to detect risky events (e.g., harsh turns).
Our findings and analysis provide valuable insights that help improve LVLM-based driving instruction systems, as illustrated in Figure~\ref{fig:overview}.

\section{Background and Related Work}

\subsection{Large-scale Vision Language Models}
Large-scale Vision Language Models (LVLMs), such as Flamingo~\cite{flamingo}, BLIP-2~\cite{blip_2}, and LLaVA~\cite{liu-etal-2023-llava}, integrate a vision encoder with a Large Language Model (LLM), enabling them to process visual inputs (e.g., images) and achieving advanced performance on Visual Question--Answering (VQA) tasks~\cite{liu-etal-2023-llava, Qwen2.5-VL}.
Visual instruction tuning is also effective for further performance improvement~\cite{liu-etal-2023-llava}.
The enhanced capabilities of LVLMs are helpful across various domains, including disease detection from clinical images~\cite{yan-etal-2024-worse,pal-sankarasubbu-2024-gemini}, artwork explanation~\cite{hayashi-etal-2024-towards, ozaki-etal-2025-towards}, and vowel prediction from MRI-based articulatory images and videos~\cite{sakajo-etal-2025-tonguescape}.

\subsection{Language Models in Driving Scenario}
LLMs and LVLMs are also helpful in driving domains, and several studies have demonstrated their capabilities~\cite{zhou-etal-vlm-ar}. 
For example, LVLMs visually understand traffic signs~\cite{Lu_Yao_Tu_Shao_Ma_Zhu_2025}, and LLMs can be utilized to develop models for processing LiDAR data~\cite{yang2023lidarllmexploringpotentiallarge} and autonomous driving~\cite{drivegpt4}.
LVLMs are also leveraged for constructing a driving dataset~\cite{covla_wacv2025}.
In the AI City Challenge~\cite{Shuo24AIC24}, LVLM-based approaches~\cite{Duan_2024_CVPR, Xuan_2024_CVPR} achieved advanced performance for the traffic safety description and analysis task, suggesting that LVLMs have the capabilities to learn and provide descriptions regarding safe driving.
However, the capabilities of LLMs to instruct safe driving with synchronized driver-facing and road-facing RGB cameras are unexplored.

\begin{table*}[t]
\centering
\small
\begin{tabularx}{\textwidth}{p{0.28\textwidth} X}
\toprule
\textbf{Primary Event} & \textbf{Description} \\
\midrule
Crash & Any visible crash involving ego-vehicle or other vehicles. \\
Forward Collision Warning & An event where the ego-vehicle is at risk of imminent front-end collision. \\
Tailgating (Following Distance) & Following another vehicle at a dangerously short distance. \\
Harsh Brake & A sudden, strong deceleration by the ego-vehicle. \\
Harsh Turn & A sharp, abrupt turn indicating potential loss of control or evasive maneuver. \\
Rolling Stop & Failure to come to a complete stop at stop signs or similar control points. \\
Mobile Usage & The driver is observed using a mobile phone (hands-on or hands-free). \\
Inattentive Driving & Observable distraction or loss of attention by the driver. \\
Lane Departure & Vehicle crosses out of its lane without clear intention or necessity. \\
Other Events & Includes seatbelt violations or distraction events. \\
\bottomrule
\end{tabularx}
\caption{Primary Event Categories for Annotation.}
\label{tab:primary-events}
\end{table*}

\begin{table*}[!t]
\centering
\small
\begin{tabularx}{\textwidth}{p{0.28\textwidth} X}
\toprule
\textbf{Sub-Event} & \textbf{Options} \\
\midrule

Lane Cut Off & Proper use of turn signal,Improper use of turn signal
 \\

Lane Change & 
Proper use of turn signal, Improper use of turn signal, 
To avoid primary event, Root cause of primary event \\

Turn (Other Vehicles) & 
Proper use of turn signal, Improper use of turn signal \\

Turn (Ego Vehicle) & 
Proper use of turn signal (listen to the audio), Improper use of turn signal \\

Signs of Aggressive Reaction & 
Vehicle maneuver, Aggressive language, Honk, None, Unknown\\

Signs of Distraction & 
Smoking, Mobile phone, Playing with hair, Drinking, Eating, 
Picking something from the floor, Reaching behind the backseat, Yawning, None, Unknown \\

Weather Condition & 
Clear, Rainy, Foggy, Snowy \\

Road Condition & 
Dry, Wet, Icy \\

Visibility Condition & 
Clear, Poor \\

Road Information & 
Highway, Highway merge, Local road, Intersection, 
3-leg intersection, School zone, Construction zone, Residential area, Rural roads, Tunnel, Pedestrian crossroad \\

Speed Management &
Decrease, Maintain, Increase \\

\bottomrule
\end{tabularx}
\caption{Sub-Event Categories and Options.}
\label{tab:annotation-events}
\end{table*}

\begin{table}[t]
    \centering
    \small
    \begin{tabular}{lrrr}
    \toprule
         & Train & Validation & Test \\
    \midrule
        Samples & 1,719 & 215 & 215 \\
        Duration (s) & 18,720 & 2,311 & 2,371 \\
        Frames & 561,223 & 69,291 & 72,836 \\
    \bottomrule
    \end{tabular}
    \caption{Dataset statistics.}
    \label{tab:dataset_stats}
\end{table}

\begin{table}[t]
    \centering
    \small
    \begin{tabular}{lr}
    \toprule
        Parameter & Value \\
    \midrule
        Batch size & 8\\
        Epoch & 3 \\
        Learning rate & 1e-5\\
        Learning rate scheduler & cosine \\
        Adam $\beta_1$ & 0.9 \\
        Adam $\beta_2$ &  0.999 \\
        Adam $\epsilon$ & 1e-8 \\
        Precision & BF16 \\
        Video Max Pixels & 16,384 \\
        Video Min Pixels & 256 \\
        Video Maxlen & 128 \\
        Video FPS & 2\\
        Seed & 42 \\
    \bottomrule
    \end{tabular}
    \caption{Hyperparameters for fine-tuning.}
    \label{tab:params_finetuning}
\end{table}

\section{Dataset Construction}
\label{sec:dataset_construction}

We constructed a dataset to evaluate LVLMs in the context of safe driving instructions, as illustrated in Figure~\ref{fig:dataset}.
The dataset comprises videos and questions, along with expert-annotated answers.

\subsection{Task}
We evaluate LVLMs' capabilities and challenges using a conversation-style VQA task related to safe driving.
We provide LVLMs with synchronized driver-facing and road-facing videos captured using RGB cameras and then ask the LVLMs to explain what happens in the video and generate instructions for safe driving.

\subsection{Video Collection}
We collected vehicle speed and video recordings from both driver-facing and road-facing RGB cameras and lined them up vertically as unified clips, placing the road-facing view on the top and the driver-facing view on the bottom.
Each video in our dataset presents synchronized views of both the driver and the road.

\subsection{Question Definition}
We adopt a Chain-of-Thought~\cite{NEURIPS2022_8bb0d291} and a conversation-style template to facilitate step-by-step reasoning.
The dataset has two questions: (1) \emph{``What is happening in the video?''} and (2) \emph{``What is recommended actions for the ego-car?''}.
We refer to the first type as \textbf{event detection} questions/answers and the second as \textbf{safe driving instruction} questions/answers.
An example of each type of question--answer pair is presented in Table~\ref{tab:qa_example}.
This Chain-of-Thought format is designed to guide the model through a reasoning process that first identifies events in the video and then infers appropriate driving actions based on those observations.

\begin{table*}[t]
    \centering
    \resizebox{\textwidth}{!}{
    \begin{tabular}{lrrrrrrrrrr}
    \toprule
    & \multicolumn{4}{c}{Event Detection} & \multicolumn{4}{c}{Safe Driving Instruction} \\
    \cmidrule(lr){2-5} \cmidrule(lr){6-9}
    \multicolumn{1}{c}{Model} & \multicolumn{1}{c}{P} & \multicolumn{1}{c}{R} & \multicolumn{1}{c}{F1} & \multicolumn{1}{c}{BLEU} & \multicolumn{1}{c}{P} & \multicolumn{1}{c}{R} & \multicolumn{1}{c}{F1} & \multicolumn{1}{c}{BLEU} \\
    \midrule
    Qwen2.5-VL-3B*\textdagger & 0.8180*\textdagger & 0.8504*\textdagger & 0.8337*\textdagger & 1.44*\textdagger & 0.8273*\textdagger & 0.8590*\textdagger & 0.8426*\textdagger & 1.66*\textdagger \\
    Qwen2.5-VL-7B & 0.8235*\textdagger & 0.8498*\textdagger & 0.8364*\textdagger & 1.26*\textdagger & 0.8086*\textdagger & 0.8662*\textdagger & 0.8364*\textdagger & 1.15*\textdagger \\
    Qwen2.5-VL-3B (FT) & 0.8901*\textdagger & 0.9041*\textdagger & 0.8970*\textdagger & 10.59* & 0.8935*\textdagger & 0.8954*\textdagger & 0.8944*\textdagger & 11.83*\textdagger \\
    Qwen2.5-VL-7B (FT) & 0.8999*\textdagger & 0.9070*\textdagger & 0.8983*\textdagger & 10.24* & 0.8994*\textdagger & 0.9002*\textdagger & 0.8997*\textdagger & 13.40*\textdagger \\
    \bottomrule
    \end{tabular}
    }
    \caption{Results of event detection and safe driving instruction tasks. Metrics include BERTScore Precision (P), Recall (R), F1, and BLEU. We performed statistical significance tests using 10,000 bootstrap resamples with a significance level of $p = 0.05$. * and \textdagger\ indicate statistically significant differences between fine-tuned and non-fine-tuned conditions and across models, respectively. The performance differences across models and between fine-tuning and non-fine-tuning conditions are statistically significant, except for the BLEU scores for the event detection between Qwen2.5-VL-3B (FT) and Qwen2.5-VL-7B (FT)
    }
    \label{tab:results}
\end{table*}
\begin{table*}[!t]
    \centering
    \resizebox{0.9\textwidth}{!}{
    \begin{tabular}{p{3.2cm}p{12cm}}
    \toprule
        Question & What is happening in the video? \\
        \midrule
        \multirow{3}{*}{Ground-truth} & The environment is icy with clear weather as you approach an intersection. A rolling stop occurs as you navigate the intersection at 8 mph while properly using your turn signal.\\
        \midrule
        \multirow{9}{*}{Qwen2.5-VL 3B} & The video shows a split-screen view of a snowy street scene and the interior of a vehicle. The top half of the screen captures the outside view, showing a snow-covered road with tire tracks, buildings on either side, and a stop sign ahead. The bottom half of the screen shows the inside of the vehicle, where a person is driving. The driver appears to be looking straight ahead, and there is a piece of paper or document on the dashboard. The scene suggests that the driver is navigating through a snowy environment, possibly preparing to make a turn or stop at the stop sign.\\
        & [BERTScore] P: 0.8202, R: 0.8554, F: 0.8374 [BLEU] 0.5540\\
        \midrule
        \multirow{10}{*}{Qwen2.5-VL 7B} & The video shows a first-person perspective from inside a vehicle, likely a truck or bus, as it drives through a snowy town. The view alternates between the road ahead and the interior of the vehicle, where a person is driving. The road appears to be partially cleared of snow, with patches of snow on the sides. There are buildings on both sides of the road, some of which have signs indicating they are businesses. The driver is focused on the road, and there are other vehicles visible in the background, including a white car and an orange vehicle parked near a building labeled ``Body Shop.'' The weather seems overcast, and the ground is covered with snow.\\
        & [BERTScore] P: 0.8139, R: 0.8439, F: 0.8286 [BLEU] 0.4693\\
        \midrule
        \multirow{4}{*}{Qwen3.5-VL 3B (FT)} & The weather is clear and the road conditions are dry as you approach an intersection. You perform a rolling stop at the intersection, maintaining a speed of 10 mph. There are no signs of distraction or aggressive reaction from your side.\\
        & [BERTScore] P: 0.8997, R: 0.9149, F: 0.9072 [BLEU] 14.3398\\
        \midrule
        \multirow{4}{*}{Qwen2.5-VL 7B (FT)} &The environment around you is snowy with icy road conditions, making it challenging to drive safely. You approach an intersection and perform a rolling stop at 10 mph while using aggressive language.\\
        & [BERTScore] P: 0.9132, R: 0.9202, F: 0.9167 [BLEU] 7.6260\\
    \bottomrule
    \end{tabular}
    }
    \caption{Sample ground-truth and generated sentences of Qwen2.5-VL 3B in event detection with evaluation metrics include BERTScore Precision (P), Recall (R), F1, and BLEU.}
    \label{tab:sample_generated_text_detection}
\end{table*}

\begin{table*}[!t]
    \centering
    \small
    \begin{tabular}{p{1.8cm}p{12cm}}
    \toprule
       Question & What is recommended actions for the ego-car?\\
       \midrule
       Ground-truth & \textbackslash ud83d\textbackslash udca1Paying extra attention to road conditions due to ice is important. Make sure to bring your vehicle to a complete stop at intersections to ensure safety, especially in icy conditions where stopping distances are longer.\\
       \midrule
       Qwen2.5-VL 3B & Based on the video description, the recommended action for the ego-car (the vehicle being driven) would be to follow the instructions given by the driver. Since the driver is looking straight ahead and appears to be focused on the road, it is likely that they have made a decision about the next move. The driver may be preparing to make a turn or stop at the stop sign. Therefore, the ego-car should follow the driver's instructions and navigate accordingly. This could involve accelerating, decelerating, or changing lanes as directed by the driver.\\
       & [BERTScore] P: 0.8356, R: 0.8400, F: 0.8378 [BLEU] 1.0798\\
       \midrule
       Qwen2.5-VL 7B & \makecell[l]{\parbox{\linewidth}{In the context of the video, the ego-car (the vehicle being driven) is navigating a snowy environment. Here are some recommended actions that could help ensure safe driving:\\\\1. **Drive Slowly**: Snowy roads can be slippery, so reducing speed helps maintain better control of the vehicle.\\\\2. **Maintain a Safe Following Distance**: Increase the distance between your vehicle and the one in front to allow more time to react to any sudden stops or changes in traffic conditions.\\\\3. **Use Low Beams**: If visibility is reduced due to snow or ice, use low beams to improve visibility without causing glare for other drivers.\\\\4. **Avoid\textbackslash u6025\textbackslash u8f6c\textbackslash u5f2f (Sharp Turns)**: Snow can cause the tires to lose traction, making sharp turns particularly dangerous. Try to avoid them if possible.\\\\5. **Check Tires**: Ensure that your tires have adequate tread depth and are properly inflated. Snow chains may also be necessary in very snowy conditions.\\\\6. **Be Aware of Other Vehicles**: Watch out for other vehicles, especially those that might be struggling to navigate the snow. Be prepared to stop or slow down if necessary.\\\\7. **Stay Alert**: Snowy conditions can be deceptive, and visibility might be reduced. Stay alert and be ready to react to any changes in the road or traffic.\\\\8. **Use Turn Signals**: Signal your intentions clearly to other drivers, especially when changing lanes or turning.\\\\9. **Keep a Full Tank of Fuel**: Snowy conditions can lead to unexpected delays, so having enough fuel is crucial.\\\\10. **Plan Your Route**: If possible, plan your route in advance to avoid areas with heavy snow accumulation or icy patches.\\\\By following these recommendations, the ego-car can navigate the snowy conditions safely and efficiently.}}\\
       & [BERTScore] P: 0.8089, R: 0.8662, F: 0.8366 [BLEU] 0.4625\\
       \midrule
       Qwen2.5-VL 3B (FT)& While it's good that you're paying attention to the road, it's important to come to a complete stop at intersections to ensure safety and compliance with traffic laws. Practice making full stops to improve your driving skills and enhance safety.\\
       & [BERTScore] P: 0.9062, R: 0.8896, F: 0.8978 [BLEU] 22.1907\\
       \midrule
       Qwen2.5-VL 7B (FT)& \textbackslash ud83d\textbackslash udca1In icy and snowy conditions, it's crucial to come to a complete stop at intersections to ensure safety. Try to remain calm and avoid aggressive language as it can distract from the primary task of driving safely.\\
       & [BERTScore] P: 0.9142, R: 0.9243, F: 0.9192 [BLEU] 22.2504\\
    \bottomrule
    \end{tabular}
    \caption{Sample ground-truth and generated sentences of Qwen2.5-VL 3 in safe drivinginstruction with evaluation metrics include BERTScore Precision (P), Recall (R), F1, and BLEU.}
    \label{tab:sample_generated_text_instruction}
\end{table*}

\subsection{Annotations}

To support structured, context-rich labeling of driving scenarios, we implemented a three-step process. 

\paragraph{Primary event selection.}
An annotator begins by selecting a single \textbf{primary event} from a predefined taxonomy of safety-critical driving behaviors, as listed in Table~\ref{tab:primary-events}. These events capture the core nature of the incident.

\paragraph{Sub-event selection.}
Next, an annotator is encouraged to select as many relevant \textbf{sub-events} as necessary to describe the contributing context. These sub-events, summarized in Table~\ref{tab:annotation-events}, include surrounding vehicle behaviors (e.g., lane changes, turn maneuvers), environmental conditions (e.g., weather, visibility), or behavioral cues (e.g., signs of aggression). This multi-label scheme enables fine-grained characterization of complex traffic scenes.

\paragraph{Summary generation.}
In the final step, a natural language annotation is automatically generated using GPT-4o~\cite{gpt4o}. The model takes as input the selected primary and sub-events, along with auxiliary data such as the vehicle’s speed at the time of the event. Based on this information, GPT-4o generates a descriptive summary that answers two key questions: ``What is happening in the video?'' and ``What is recommended actions for the ego-car?''
Finally, experts manually review the generated descriptions and confirm the quality.

\subsection{Data Statistics}
Table~\ref{tab:dataset_stats} shows our dataset statistics.
A video has an approximate duration of 10 seconds and 30 frames per second.
The number of primary events and sub-event options is provided in Appendix~\ref{sec:dataset_stats_detail}.

\section{Experimental Settings}

\subsection{Dataset}
We use our dataset introduced in Section~\ref{sec:dataset_construction} and treat the videos as two frames per second.

\begin{table*}[t]
    \centering
    \small
    \begin{tabular}{p{0.23\linewidth}|p{0.7\linewidth}}
\toprule
Model \& Event Type &  Top-10 4-gram words \\
\midrule
Qwen2.5-VL-3B & \\
Event Detection & half of the video; The video shows a; top half of the; The video shows two; the interior of the; The top half of; bottom half of the; The bottom half of; The interior of the; interior of the vehicle. \\
Safe Driving Instuction & Based on the video; on the video description,; the ego-car (the vehicle; for the ego-car (the; there are no specific; are no specific actions; recommended actions for the; the video description, there; video description, there are; a safe distance from. \\
\hline
Qwen2.5-VL-3B (FT) & \\
Event Detection & no signs of distraction; signs of distraction or; of distraction or aggressive; There are no signs; are no signs of; The footage shows you; footage shows you driving; distraction or aggressive reaction; or aggressive reaction from; aggressive reaction from your. \\
Safe Driving Instruction & While it's good that; come to a complete; to a complete stop; a complete stop at; it's good that you're; to come to a; it's important to come; important to come to; complete stop at intersections; stop at intersections to. \\
\hline
Qwen2.5-VL-7B & \\
Event Detection & The video appears to; half of the screen; of the screen shows; the screen shows the; video appears to be; the interior of the; appears to be a; shows the interior of; to be a split-screen; be a split-screen view. \\
Safe Driving Instruction & In the context of; the context of the; are some general recommendations; some general recommendations for; the ego-car (the vehicle; context of the video,; a safe distance from; ego-car (the vehicle being; distance from the vehicle; for the ego-car (the. \\
\hline
Qwen2.5-VL-7B (FT) & \\
Event Detection & signs of distraction or; of distraction or aggressive; no signs of distraction; There are no signs; are no signs of; clear weather and dry; weather and dry road; The footage shows you; footage shows you driving; and dry road conditions. \\
Safe Driving Instruction & come to a complete; to a complete stop; a complete stop at; increase your following distance; maintain a safe following; your following distance to; to maintain a safe; a safe following distance; important to maintain a; safe following distance to. \\
\bottomrule
\end{tabular}
    \caption{The top 10 4-grams in each response.}
    \label{tab:top10_n_grams}
\end{table*}

\subsection{Models}
In this study, we utilize Qwen2.5-VL~\cite{Qwen2.5-VL} 3B and 7B models, which are available even with limited computational resources.
We also fine-tune these models on our dataset.
We refer to fine-tuned Qwen2.5-VL-3B and fine-tuned Qwen2.5-VL-7B as Qwen2.5-VL-3B~(FT) and Qwen2.5-VL-7B~(FT), respectively.

\subsection{Training and Inference}
Models receive instruction and video inputs, while auxiliary sensor data, e.g., vehicle speed, was incorporated during dataset construction. This approach reflects the practical consideration that dashcams are easily deployable, whereas sensor installation requires additional costs.

\paragraph{Training.}
We freeze the vision encoder and train only the language model with full-parameter supervised fine-tuning.
LVLMs fine-tuning is conducted using the LLaMA-Factory~\cite{zheng-etal-2024-llamafactory} with DeepSpeed ZeRO stage 2~\cite{zero}.
We fine-tuned LVLMs using eight NVIDIA A100-SXM4-40GB GPUs and used LLaMA-Factory version 0.9.2.dev0 with minor modifications to load models correctly.
Table~\ref{tab:params_finetuning} provides the hyperparameters.

\paragraph{Inference.}
We test LVLMs and fine-tuned LVLMs under the zero-shot setting.
The evaluations are performed on an NVIDIA L4 GPU.

\subsection{Metrics}
We evaluate the quality of generated text by comparing it to the reference text in the dataset using BERTScore~\cite{bert-score} and BLEU socres~\cite{papineni-etal-2002-bleu} as evaluation metrics.
We use the original implementation\footnote{{\url{https://github.com/Tiiiger/bert_score}}} for BERTScore using RoBERTa~\cite{liu2019robertarobustlyoptimizedbert} and sacreBLEU~\cite{post-2018-call}\footnote{\url{https://github.com/mjpost/sacrebleu}} for BLEU scores.

\section{Result and Discussion}
\label{sec:result_discussion}

Table~\ref{tab:results} shows the results of each model on our dataset, and Tables~\ref{tab:sample_generated_text_detection} and \ref{tab:sample_generated_text_instruction} show the samples of generated text for event detection and safe driving instruction.
Before fine-tuning, Qwen2.5-VL-3B performs better in terms of F1 score on BERTScore for the safe driving instruction than Qwen2.5-VL-7B, while the 7B model outperforms the 3B model in event detection.
This suggests that the parameter size is irrelevant to the task performance of pre-trained models.
Fine-tuning improves both BERT scores and BLEU scores, indicating that models can learn this task correctly.

\subsection{Result Comparison}
\label{sec:result_comparison}

\begin{table}[t]
    \centering
    \small
    \begin{tabular}{lrr}
    \toprule
         & \makecell{ Detection} & \makecell{ Instruction} \\
    \midrule
        Qwen2.5-VL-3B & 84.4359 & 82.8134 \\
        Qwen2.5-VL-7B & 83.6289 & 85.7856 \\
        Qwen2.5-VL-3B (FT) & 96.4625 & 95.8846 \\
        Qwen2.5-VL-7B (FT) & 95.5442 & 93.5108 \\
    \bottomrule
    \end{tabular}
    \caption{Self-BLEU scores. ``Detection'' and ``Instruction'' denote ``Event Detection'' and ``Safe Driving Instruction'', respectively.}
    \label{tab:result_self_bleu}
\end{table}

\paragraph{Parameter size influence.}
The difference in the F1 score of BERTScore between Qwen2.5-VL-3B and Qwen2.5-VL-7B on the event detection was not significant, indicating that the performance of the event detection task is saturated in the zero-shot setting.
The results demonstrate that the 3B model performs better for safe driving instruction than the 7B model, which can be attributed to the output tendencies in the 7B model.
The 7B model outputs general recommendations for safe driving instruction before fine-tuning, as shown in Table~\ref{tab:top10_n_grams}, resulting in lower precision and higher recall.
For further analysis, we also computed self-BLEU scores~\cite{self_bleu} for each event using the outputs generated by each model to assess diversity.
The self-BLEU scores, as shown in Table~\ref{tab:result_self_bleu}, also indicate that the 7B model outputs less diverse texts for the safe driving instruction when compared with the 3B model.

\begin{figure*}[t]
    \centering
    \includegraphics[width=0.8\linewidth]{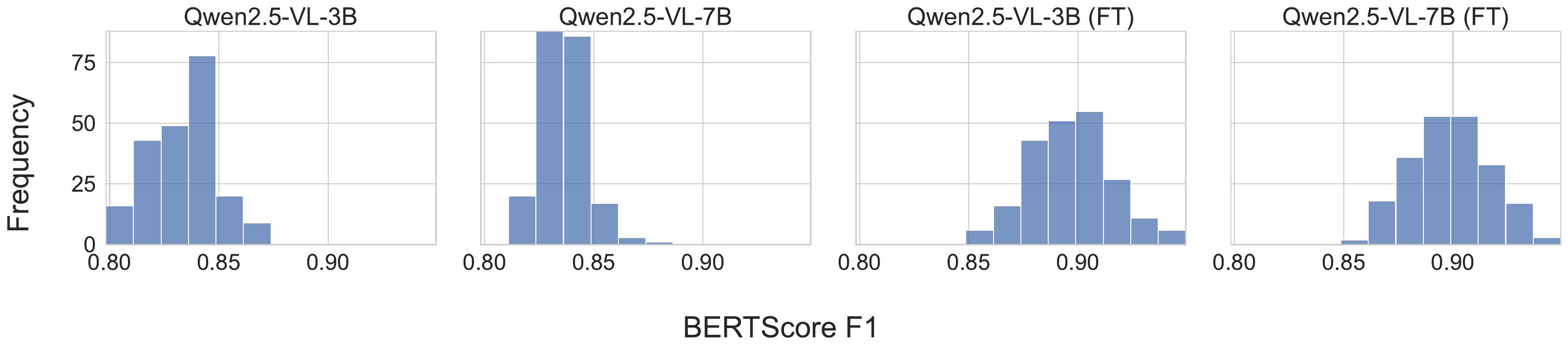}
    \caption{Score distribution of BERTScore F1 of event detection.}
    \label{fig:distribution_f1_event_detection}
\end{figure*}

\begin{figure*}[t]
    \centering
    \includegraphics[width=0.8\linewidth]{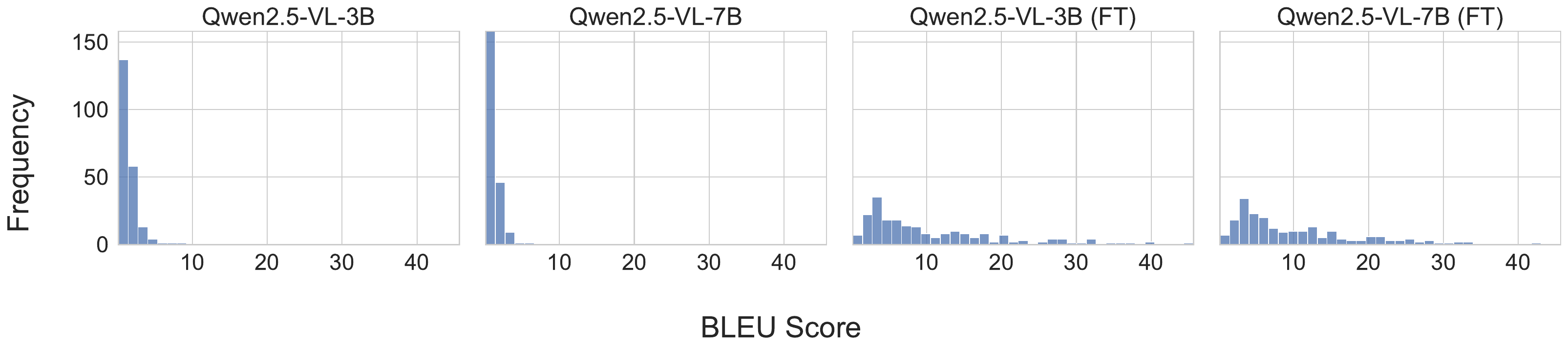}
    \caption{Score distribution of BLEU of event detection.}
    \label{fig:distribution_bleu_event_detection}
\end{figure*}

\begin{figure*}[!t]
    \centering
    \includegraphics[width=0.8\linewidth]{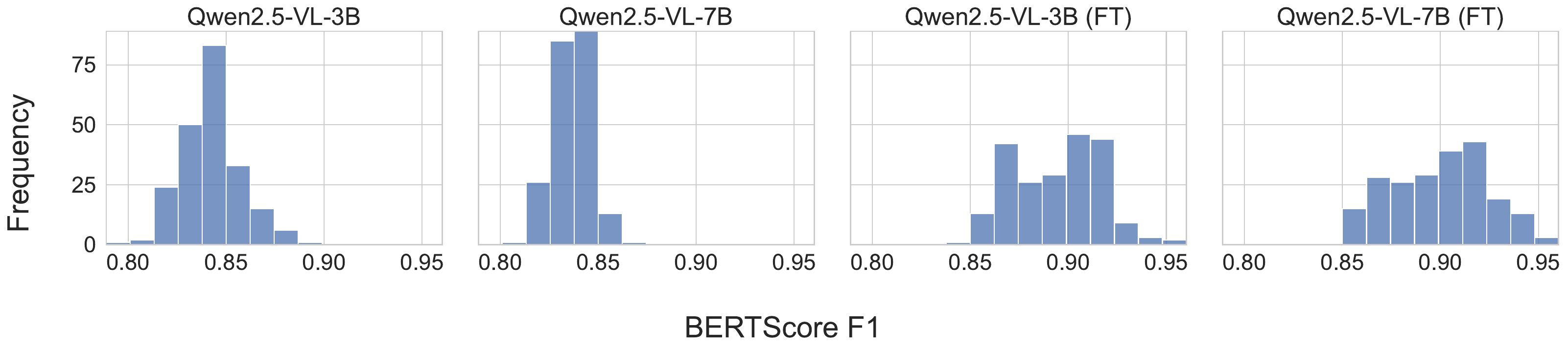}
    \caption{Score distribution of BERTScore F1 of safe driving instruction.}
    \label{fig:distribution_f1_instruction}
\end{figure*}

\begin{figure*}[!ht]
    \centering
    \includegraphics[width=0.8\linewidth]{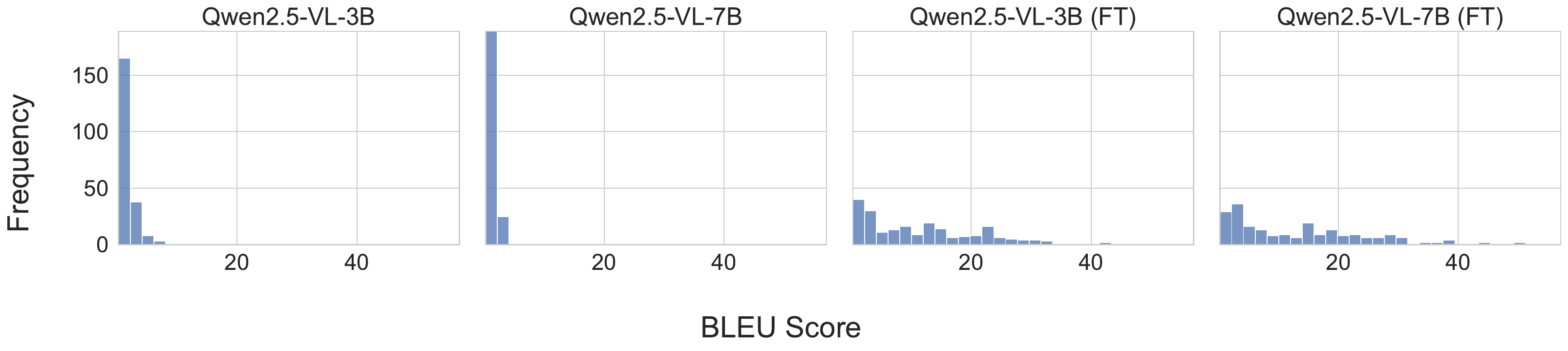}
    \caption{Score distribution of BLEU of safe driving instruction.}
    \label{fig:distribution_bleu_instruction}
\end{figure*}

\paragraph{Performance improvement by fine-tuning.}
Fine-tuning improves overall performance, and Qwen2.5-VL-7B (FT) outperforms Qwen2.5-VL-3B (FT) on both tasks in terms of BERTScore, while Qwen2.5-VL-3B outperforms Qwen2.5-VL-7B before fine-tuning.
Figures \ref{fig:distribution_f1_event_detection}, \ref{fig:distribution_bleu_event_detection} \ref{fig:distribution_f1_instruction} and \ref{fig:distribution_bleu_instruction} also show that fine-tuning improves overall performance.
On the other hand, Table~\ref{tab:results} shows that the difference in the BLEU scores between both fine-tuned models on the event detection task is not significant.
This suggests that a larger parameter size has a positive effect on the fine-tuning performance of LVLMs for this task.
In contrast, the final performance after fine-tuning remains consistent across model sizes with respect to BLEU scores.


\subsection{Error Analysis}
\label{sec:error_analysis}
We focus on the subset of samples for which BERTScore of the safe driving instruction falls within the bottom 25\%. 
Approximately 4\% of all the samples are shared across the bottom 25\% subsets for all models, which we refer to as the ``difficult subset''.
Within this subset, 33\% of the samples are annotated as good driving, and another 33\% involve scenarios where the ego-car is turning right and left.
Although all models generated recommendations to improve already good driving behaviors, the suggestions for safer driving varied slightly, resulting in relatively low scores.

For the turning right and left scenarios, the gold answers typically recommend turning while reducing speed.
However, even fine-tuned models produced irrelevant suggestions, such as mentioning a stop sign not presented in the video.
These observations suggest that while fine-tuned models are capable of generating various safety-related suggestions, they still struggle to detect issues such as excessive speed during turns.

In the difficult subset, approximately 10\% have errors related to the driver-facing view, where a driver holds and uses a phone while keeping their eyes on the road.
This might suggest that LVLMs can provide safe driving instruction regarding drivers' behaviors, while they struggle to generate it regarding vehicle behaviors.
This phenomenon is explained by the relative ease with which LVLMs can detect a driver holding an object, as opposed to estimating vehicle speed, which requires more temporal reasoning.

\subsection{Unimodal Biases}
\label{sec:unimodal_biases}
As discussed in Section~\ref{sec:result_comparison}, in several cases, pre-trained models provide general suggestions regardless of the videos, as shown in Table~\ref{tab:sample_generated_text_instruction}.
We also observed that the fine-tuned models mentioned objects not presented in the video in Section~\ref{sec:error_analysis}.
This behavior can be attributed to unimodal biases, specifically language biases~\cite{Goyal_2017_CVPR, Agrawal_2018_CVPR, Zhu2020, 9156448, chen-etal-2024-quantifying}, where models' outputs are biased toward textual information in the given inputs.
However, while language biases have been attributed to the model's learning of the relationships between question-answer pairs in the training data~\cite{Agrawal_2018_CVPR}, it is unclear whether the training data includes question-answer pairs regarding safe driving instruction.
Given that \citet{sakajo-etal-2025-tonguescape} observed the model generating identical answers for different images with the same question in a phonetics-related VQA task, our findings suggest that language priors emerge in certain domains.

\subsection{Task Difficulty and Application Possibility}
\label{sec:task_difficulity}
The results reveal that this task is challenging for LVLMs without fine-tuning, whereas fine-tuning improves performance.
Our error analysis in Section~\ref{sec:error_analysis} also indicates that several failure cases happen for good driving videos, and suggestions for safer driving vary slightly.
Those discussions suggest that our fine-tuned models can be applied to safe driving instruction systems, although several challenges remain in certain situations, such as instructing against a harsh turn.

\section{Conclusion}
In this study, we constructed a dataset comprising synchronized driver-facing and road-facing video streams, along with step-by-step question--answer pairs.
We fine-tuned LVLMs on our dataset for safe driving instructions and investigated their capabilities and performance in detecting risky events and providing safe driving instructions.
Our experimental results reveal that fine-tuned LVLMs demonstrate the capability of suggesting safety-aware driving instructions, while detecting several events remains challenging, even for fine-tuned models.
Our findings suggest that LVLMs can be safe driving instructors, although there is room for improvement.

\section*{Limitations}
\paragraph{Dataset size.}
As described in Section~\ref{sec:dataset_construction}, our dataset comprises 1,719 training samples, 215 validation samples, and 215 test samples, which can be considered relatively small in scale.
However, the collection of synchronized driver-facing and road-facing views requires a complicated setup, characterizing this task as a low-resource scenario.
In this study, we investigated model performance using the current dataset as an initial step, with evaluation on a larger dataset left for future work.

\paragraph{Dataset quality.}
The instructions in our dataset were generated using GPT-4o, which might raise concerns regarding their quality.
However, as detailed in Section~\ref{sec:dataset_construction}, the generated texts were manually reviewed to ensure the quality.

\paragraph{Model selection.}
In this study, we selected two base models: Qwen2.5-VL-3B and Qwen2.5-VL-7B.
While this choice might constrain our investigation of the scaling law in this task and performance variation across models, it remains justifiable.
The Qwen2.5-VL series achieves advanced performance on various benchmarks, including Video-MME~\cite{fu2024video}, and our objective is to evaluate model effectiveness for driving instruction.
Accordingly, focusing on the Qwen2.5-VL series and its relatively small variants is appropriate for our investigation.

\section*{Ethical Considerations}
Our dataset contains videos that capture drivers. We collect these videos legitimately and use them within the prescribed scope.

\section*{Acknowledgement}
We thank the anonymous reviewers for their valuable comments and suggestions.
This project is supported by Teatis inc. through the provision of the dataset and computational resources.

\bibliography{anthology, custom}

\appendix

\section{Dataset Statistics (Detail)}
\label{sec:dataset_stats_detail}

Table~\ref{tab:num_event_dataset} shows the number of events or options in each dataset split.

\small
  \bottomcaption{The number of each event or option in each dataset split. Val. denotes the validation set.}
  \label{tab:num_event_dataset}
  \begin{supertabular}{lrrr}
  \toprule
       & Train & Val. & Test  \\
  \midrule
    \multicolumn{4}{l}{\textbf{Primary Events}}\\
    \quad Crash & 3 & 0 & 0 \\
    \quad Forward Collision Warning & 41 & 6 & 7 \\
    \quad Tailgating & 90 & 21 & 7 \\
    \quad Harsh Brake & 253 & 20 & 26 \\
    \quad Harsh Turn & 15 & 4 & 2 \\
    \quad Rolling Stop & 308 & 42 & 42 \\
    \quad Mobile Usage & 87 & 14 & 11 \\
    \quad Inattentive Driving & 142 & 17 & 20 \\
    \quad Lane Departure & 0 & 0 & 0 \\
    \multicolumn{4}{l}{\textbf{Sub Events}}\\
    Lane Cut Off & & & \\
    \quad Improper use of turn signal & 18 & 3 & 0 \\
    \quad Proper use of turn signal & 21 & 7 & 1 \\
    Lane Change & & & \\
    \quad Improper use of turn signal & 11 & 0 & 1 \\
    \quad Proper use of turn signal & 84 & 11 & 9 \\
    \quad To avoid primary event & 21 & 2 & 3 \\
    \quad Root cause of primary event & 24 & 3 & 2 \\
    Turn (Other Vehicles) & & & \\
    \quad Proper use of turn signal & 10 & 1 & 3 \\
    \quad Improper use of turn signal & 3 & 0 & 0 \\
    Turn (Ego Vehicle) & & & \\
    \quad Proper use of turn signal & 114 & 15 & 20 \\
    \quad Improper use of turn signal & 102 & 18 & 10 \\
    Signs of Aggressive Reaction & & & \\
    \quad Vehicle maneuver & 1 & 0 & 0 \\
    \quad Aggressive language & 18 & 1 & 2 \\
    \quad Honk & 9 & 2 & 0 \\
    \quad None & 634 & 72 & 85 \\
    \quad Unknown (Dashcam Issue) & 209 & 30 & 25 \\
    Signs of Distraction & & & \\
    \quad Smoking & 26 & 3 & 5 \\
    \quad Mobile phone & 18 & 3 & 2 \\
    \quad Playing with hair & 2 & 1 & 0 \\
    \quad Drinking & 14 & 0 & 2 \\
    \quad Eating & 18 & 2 & 3 \\
    \quad Picking something from floor & 2 & 0 & 1 \\
    \quad Reaching behind backseat & 0 & 1 & 0 \\
    \quad Yawning & 1 & 0 & 0 \\
    \quad None & 580 & 63 & 75 \\
    \quad Unknown & 218 & 32 & 25 \\
    Weather Condition & & & \\
    \quad Clear & 1380 & 184 & 172 \\
    \quad Rainy & 92 & 6 & 10 \\
    \quad Foggy & 0 & 1 & 0 \\
    \quad Snowy & 60 & 7 & 9 \\
    Road Condition & & & \\
    \quad Dry & 1362 & 181 & 170 \\
    \quad Wet & 98 & 6 & 11 \\
    \quad Icy & 72 & 11 & 11 \\
    Visibility Condition & & & \\
    \quad Clear & 0 & 0 & 0 \\
    \quad Poor & 0 & 0 & 0 \\
    Road Information & & & \\
    \quad Highway & 893 & 124 & 102 \\
    \quad Highway merge & 22 & 3 & 2 \\
    \quad Local Road & 8 & 0 & 0 \\
    \quad Intersection & 348 & 28 & 46 \\
    \quad 3-Leg intersection & 166 & 31 & 26 \\
    \quad School zone & 0 & 1 & 0 \\
    \quad Construction Zone & 10 & 3 & 2 \\
    \quad Residential area & 66 & 8 & 14 \\
    \quad Rural roads & 12 & 2 & 1 \\
    \quad Tunnel & 1 & 0 & 0 \\
    \quad Pedestrian crossroad & 22 & 2 & 1 \\
    \quad Parking & 27 & 2 & 6 \\
    Speed Management & & & \\
    \quad Decrease & 37 & 5 & 1 \\
    \quad Maintain & 409 & 59 & 48 \\
    \quad Increase & 43 & 9 & 1 \\
  \bottomrule
  \end{supertabular}
\normalsize

\end{document}